\documentclass[conference,a4paper]{IEEEtran}

\usepackage[final]{graphicx}
\usepackage[reqno]{amsmath}
\usepackage{amssymb}
\usepackage{amsmath}
\usepackage{subfig}
\usepackage{epstopdf}
\usepackage{xcolor}
\usepackage{float}
\usepackage[paper=letterpaper,tmargin=1in,bmargin=.8in,left=.8in, right=.8in]{geometry}
\usepackage{comment}

\begin{document}

\sloppy
 
\title{Combatting Adversarial Attacks through Denoising and Dimensionality Reduction:\\ A Cascaded Autoencoder Approach}
\author{\IEEEauthorblockN{Rajeev Sahay, Rehana Mahfuz, and Aly El Gamal}
 \IEEEauthorblockA{School of Electrical and Computer Engineering \\Purdue University\\ Email: \{sahayr, rmahfuz, elgamala\}@purdue.edu}
\thanks{Rajeev and Rehana have equally contributed to this work.}

}
\maketitle

\begin{abstract}
Machine Learning models are vulnerable to adversarial attacks that rely on perturbing the input data. This work proposes a novel strategy using Autoencoder Deep Neural Networks to defend a machine learning model against two gradient-based attacks: The Fast Gradient Sign attack and Fast Gradient attack. First we use an autoencoder to denoise the test data, which is trained with both clean and corrupted data. Then, we reduce the dimension of the denoised data using the hidden layer representation of another autoencoder. We perform this experiment for multiple values of the bound of adversarial perturbations, and consider different numbers of reduced dimensions. When the test data is preprocessed using this cascaded pipeline, the tested deep neural network classifier yields a much higher accuracy, thus mitigating the effect of the adversarial perturbation.
\end{abstract}
\section{Introduction}
State of the art machine learning algorithms have revolutionized automated classification technologies in various fields like computer vision, natural language processing, and biometric information security \cite{CV} \cite{NLP} \cite{biometric}. High classification accuracy in the foregoing applications has led to the deployment of learning algorithms in a multitude of environments. Yet, recent studies have exposed the vulnerabilities of accurate machine learning classification in the presence of adversarial attacks \cite{adv}. More specifically, the injection of visually imperceptible $l_2$ and $l_\infty$ bounded perturbations into the input testing data has shown to render even the most robust classifiers useless. For example, both Support Vector Machines (SVMs) and Artificial Neural Networks with robust classification metrics have resulted in nearly $0\%$ accuracy when processing perturbed data \cite{sparsitysvm} \cite{lintrans}. This paper presents novel defense strategies, which are not only capable of combatting such adversarial attacks, but are also more robust than current standards in the literature.

In this study, we consider an attacker-defender scenario. The attacker aims to optimally perturb input data such that it causes a trained multi-layer fully-connected artificial neural network to misclassify the input sample. The defender employs various defense strategies to resist the misclassification caused by the attacker. We consider three defense strategies: a stand-alone Denoising Autoencoder (DAE), a reduced-dimension representation of the input obtained from a fully-connected autoencoder neural network, and finally a cascade of the aforementioned defenses connected in series. Each defense strategy is tested in both a semi-white box environment, in which the adversary has full knowledge of the trained neural network classifier architecture (but not the parameter initialization values), and a black-box environment in which the adversary has no prior knowledge of the trained classifier architecture. In either scenario, the attacker is blind to the pre-processing defense algorithms. Finally, we assume that the attacker employs an untargeted misclassification attack. In other words, the attacker's sole objective is misclassification, without particular emphasis on a specific output.

\subsection{Related Work}
Several attacks that rely on effective perturbation of the input samples exist in the literature, such as the DeepFool algorithm \cite{deepfool}, the Carlini and Wagner $l_2$ attack \cite{carlini}, and the Fast Gradient Sign (FGS) algorithm \cite{fgs}. Each attack relies on perturbing the input sample slightly differently by optimizing different network parameters. Previous studies have investigated the effectiveness of using sparsity as a defense against such adversarial perturbations. For example, Z. Marzi et al. \cite{sparsitysvm} shows that enforcing sparsity by taking the $K$ out of $N$ largest elements in magnitude, and zeroing the rest, for an inputted classifier sample in the wavelet domain, successfully decreases the misclassification caused by an adversarial attack on a Support Vector Machine (SVM). Furthermore, A. N. Bhagoji et al. \cite{lintrans} shows that projecting high dimensional data onto a lower dimensional subspace using Principle Component Analysis (PCA) is effective in decreasing adversarial success. 
Further, A. N. Bhagoji et al. \cite{lintrans} extend their defense strategies beyond SVMs and successfully demonstrate PCA as a successful defense in multi-class scenarios on artificial neural networks. In this paper, we present three distinct defense strategies to combat adversarial attacks on deep neural networks that result in more robust defenses than previous works have achieved.

Furthermore, S. Gu et al. \cite{towardsdnn} showed that DAEs can be used to effectively remove Gaussian noise injected into input samples, but it does not explore their application in the scenario of an adversarial attack. However, S. Gu et al. \cite{towardsdnn} does reveal that multi-layer denoising autoencoder architectures can be used to effectively eliminate injected noise. We aim to extend the use of DAEs in different scenarios so that they can be used as a defense against perturbed inputs. Specifically, we train a DAE architecture capable of outputting clean samples regardless of whether the inputted data was benign or corrupted.  

\subsection{Contributions}
We present three novel defense strategies to combat adversarial machine learning attacks: The Denoising Autoencoder (DAE), dimensionality reduction using the learned hidden layer of a fully-connected autoencoder neural network, and a cascade of the DAE followed by the learned reduced dimensional subspace in series. Each of our defense strategies are used as pre-processing defense mechanisms which aim to reduce the effect of the adversary on a trained fully-connected multi-layer artificial neural network. We show that the DAE alone is an effective defense capable of increasing the average accuracy across a noise range of $[0.00, 0.50]$ from $13.76\%$ to $95.55\%$ in a semi-white box environment, and from $18.54\%$ to $78.37\%$ in a black-box environment under an $l_\infty$ bounded attack. Furthermore, we show that dimensionality reduction using an autoencoder is also an effective defense against adversarial attacks, that is capable of increasing the average accuracy across a noise range of $[0.00, 0.50]$ from $13.76\%$ to $51.09\%$ in a semi-white box environment, and from $18.54\%$ to $48.62\%$ in a black box environment under an $l_\infty$ bounded attack. Finally, our \textbf{novel cascaded architecture} increases the average accuracy across the same noise range from $13.76\%$ to $96.27\%$ in a semi-white box environment, and from $18.54\%$ to $79.88\%$ in a black box environment under an $l_\infty$ bounded attack. We also obtain similar results for attacks induced according to the $l_2$ norm.\footnote{Code is available at https://github.com/rajeevsahay/ae-defenses} \textbf{To the best of our knowledge, the gains in accuracy achieved in this study have not been achieved by any other defense strategies against adversarial attacks for neural network classifiers}.
\section{Problem Description} 
\subsection{Adversarial Attacks on Neural Networks}\label{subsec:aana}
Neural networks are vulnerable to different types of adversarial attacks aimed at causing misclassification. Attacks which corrupt training data are called poisoning attacks \cite{Poisoning}, resulting in an incorrectly trained classifier. Attacks which corrupt test data are called evasion attacks. While several methods exist to optimally perturb neural network inputs, this paper explores two gradient-based evasion attack methods: the Fast Gradient Sign (FGS) attack and the Fast Gradient (FG) attack. The FGS attack \cite{FGS-proposal} adds an $l_\infty$-bounded perturbation $\eta$ such that $\eta = ||x-\widetilde{x}||_\infty$,  where $\it{x}$ and $\widetilde{x}$ correspond to the original and corrupted sample, respectively. This results in the following corruption: 
\begin{equation}\label{eq:fgs} 
    \widetilde{x}(\eta) = x + \eta * sign(\triangledown_x J(\theta,x,y)),
\end{equation}
where $\widetilde{x}(\eta)$ is the perturbed version of the input data sample $x$, whose predicted class is $y$, and $J(\theta,x,y)$ represents the cost function used to train the neural network, where $\theta$ represents the network weights. The sign of the gradient of this cost function is scaled by $\eta$  and added to the benign data sample to perturb it. A moderately low choice of $\eta$, below an approximate threshold of 0.15, results in visually imperceptible perturbations. We consider a range of $\eta$ between 0.0 and 0.5.

A variation of the FGS attack is the Fast Gradient (FG) attack, which introduces an $l_2$ bounded error such that $\epsilon = ||x-\widetilde{x}||_2$. This results in the corrupted sample as shown in (2).

\begin{equation}\label{eq:fg}
    \widetilde{x}(\epsilon) = x + \epsilon * \dfrac{ \triangledown_x J(\theta,x,y)}{||\triangledown_x J(\theta,x,y)||_2}
\end{equation}
For $l_2$-bounded perturbations, the perturbations are imperceptible for larger bounds as well. We consider a range of $\epsilon$ between 0.0 and 3.5.

There are different categories of attacks, based on how much knowledge the adversary has about the trained model and the defense mechanism. A white box scenario is when the adversary has access to the trained model and the defense strategy. It is highly unlikely for an adversary to have so much information about the classifier, which is why we omit this scenario in our experiments. A slightly more realistic scenario is the semi-white box setting, in which the adversary has access to the training dataset and classifier architecture, but is blind to the defense mechanism. An even more practical setting is the black-box setting, where the attacker has access to the training data only, and arbitrarily (or using imperfect knowledge) selects a classifier architecture, based on which it generates its adversarial examples. The black box scenario represents the most realistic attacker-defender situation and illustrates the application of our adversarial defense algorithms in defender-attacker mismatched classifier scenarios. 
\subsection{Experimental Setup}\label{subsec:es}
We evaluate the effect of our defenses on a neural network classifier trained on the MNIST data set \cite{mnist}, which consists of 60,00 training examples and 10,000 testing examples of black-and-white handwritten digits centered and normalized to a of size 28 x 28 pixels, where each image represents a digit from 0 to 9. Furthermore, each pixel value is normalized to lie in [0, 1]. We consider a fully connected neural network classifier with an input layer to accept the 784 pixel values followed by two layers of 100 neurons each. The output layer contains 10 neurons, which correspond to one of ten possible digit classes. Throughout this paper, this model is referred to as the FC-100-100-10 architecture. In the black box scenario, the adversary generates perturbations based on an FC-200-200-100-10 architecture, whose name corresponds to the same naming convention as that of the FC-100-100-10 model. Each layer has ReLu activation functions, except the last layer, which has a softmax activation function. The batch size used for training is 200, the optimizer is adam, the loss function is categorical crossentropy, and 100 epochs are used to train both the classifier and the autoencoders. It is important to note that distinct FC-100-100-10 architectures were trained for each of the four attack environments, which results in slightly different baseline accuracy values due to unique parameter initilizations in each scenario. All neural networks were implemented using the Keras library \cite{keras} in Python, and adversarial perturbations are generated using the Cleverhans library \cite{fgs}.
\subsection{Defense Strategies}\label{subsec:def}
Because adversarial perturbations decrease classification accuracy, a variety of defense strategies have been proposed to combat this effect, such as network distillation \cite{distillation} and adversarial retraining \cite{FGS-proposal}.
Here, we process the test data using autoencoders before classifying it, with an aim to recover the performance degradation due to perturbations. Autoencoders can be used to denoise and learn compact representations of data. An autoencoder is a neural network that attempts to reproduce an output, which is approximately equivalent to the input through learned encoder and decoder functions.

A traditional autoencoder learns the underlying manifold of the training data's distribution, which is used to reconstruct the input at the output. This manifold can be used advantageously by training the DAE with benign and corrupted samples that are mapped only to clean samples. As a result, the autoencoder will learn an underlying vector field that points in the direction of the manifold in which the clean samples lie. Thus, upon the introduction of a perturbation, the magnitude of each arrow in the vector field will indicate the direction in which the data must be moved to map the sample to its clean representation. After training the DAE using both uncorrupted and corrupted data, the output of the propagated data will consist only of clean samples.

Furthermore, when a hidden layer of an autoencoder contains lesser dimensions than the input sample, we can also use it to learn a compressed representation of the input. This is achieved by constructing a bottleneck architecture in which the input and output have an identical number of units, but the hidden layer consists of the number of units corresponding to the desired reduced dimension. The output of the hidden layer of the trained autoencoder is then extracted and used as the input into the classifier.  
\subsubsection{Denoising using Autoencoders}\label{subsubsec:dua}
Propagating the data through an autoencoder forces the network to learn the underlying manifold of the input by mapping noisy data back onto the clean manifold distribution of the data. Therefore, the output of the Denoising Autoencoder (DAE) is an uncorrupted version of the input sample. However, to generate such a model, the training dataset must consist of both clean and corrupted samples and their corresponding clean, uncorrupted ideal outputs. For both considered semi-white box and black box attack environments, the defender simulates adversarial training data, using the FGS attack, by adding perturbations, with a magnitude of $\eta = 0.25$, to the training samples based on the gradient of the FC-100-100-10 model. The DAE is then trained with both clean and perturbed data. The architecture of the denoising autoencoder was chosen to be a fully connected neural network with the following architecture: 784-256-128-64-128-256-784. The DAE is then trained with 60,000 clean data samples, and 60,000 corrupted data samples, and optimized using the mean squared error cost function over 150 epochs using a batch size of 256. The DAE's mean squared error reaches a minimum at 0.0049, indicating the network's robust ability to approximate the uncorrupted input of both clean and perturbed samples. Our experiments revealed that the DAE trained to combat $l_\infty$ attacks delivers a more robust defense, compared to using a DAE trained to combat $l_2$ attacks, even when combatting $l_2$ bounded attacks. Thus, all FG attacks were defended using the DAE trained to combat $l_\infty$ attacks. 
\subsubsection{Dimensionality Reduction using Autoencoders}\label{subsubsec:drua}
The output of the bottleneck layer of an autoencoder gives a compressed representation of the data. We experiment with different numbers of hidden layers to compress the initial 784 input dimensions to a smaller number of dimensions $k$. The various values of $k$ considered are 331, 100, 80, 60, 40 and 20, with different magnitudes of the perturbation bound to see which compressed representation gives the best accuracy. Instead of using all 784 input features, we use a lesser number of features as the input data, both for training and for testing. This also means that the size of the input layer changes, which means that it is unnecessary to train it with the architecture that uses 784 input dimensions. For each value of $k$, the autoencoder architecture used was 784-$k$-784. Each model was a fully connected neural network trained with the MNIST training set of 60,000 samples.

\subsubsection{Cascaded Defense}\label{subsubsec:cs}
We consider a novel defense strategy, in which we cascade the DAE with dimensionality reduction. First, we denoise the test data using the denoising autoencoder, after which we reduce its dimensionality using the hidden layer representation of an autoencoder, before sending it as an input into the classifier. Both methods are implemented in series identically to their independent workings. 

\section{Results}
In this section, we show the efficacy of each pre-processing defense mechanism used in this study in four distinct attack environments: the semi-white box FGS attack, the semi-white box FG attack, the black box FGS attack, and the black box FG attack. For each of the four attack environments, we first show the accuracy of classifying the MNIST testing set plotted against the perturbation magnitude with no defensive mechanism. We then show the effectiveness of the DAE as a stand-alone defense, followed by the effectiveness of dimensionality reduction as a stand-alone defense. Finally, the results produced from cascading the DAE with dimensionality reduction for use as a single defense is shown for different numbers of reduced dimensions $k$, where $k \in \{20, 40, 60, 80, 100, 331\}$.
\subsection{Semi-White Box FGS Attack}\label{sec:swbfgs}
\begin{figure}[htb]
	\includegraphics[width=\linewidth]{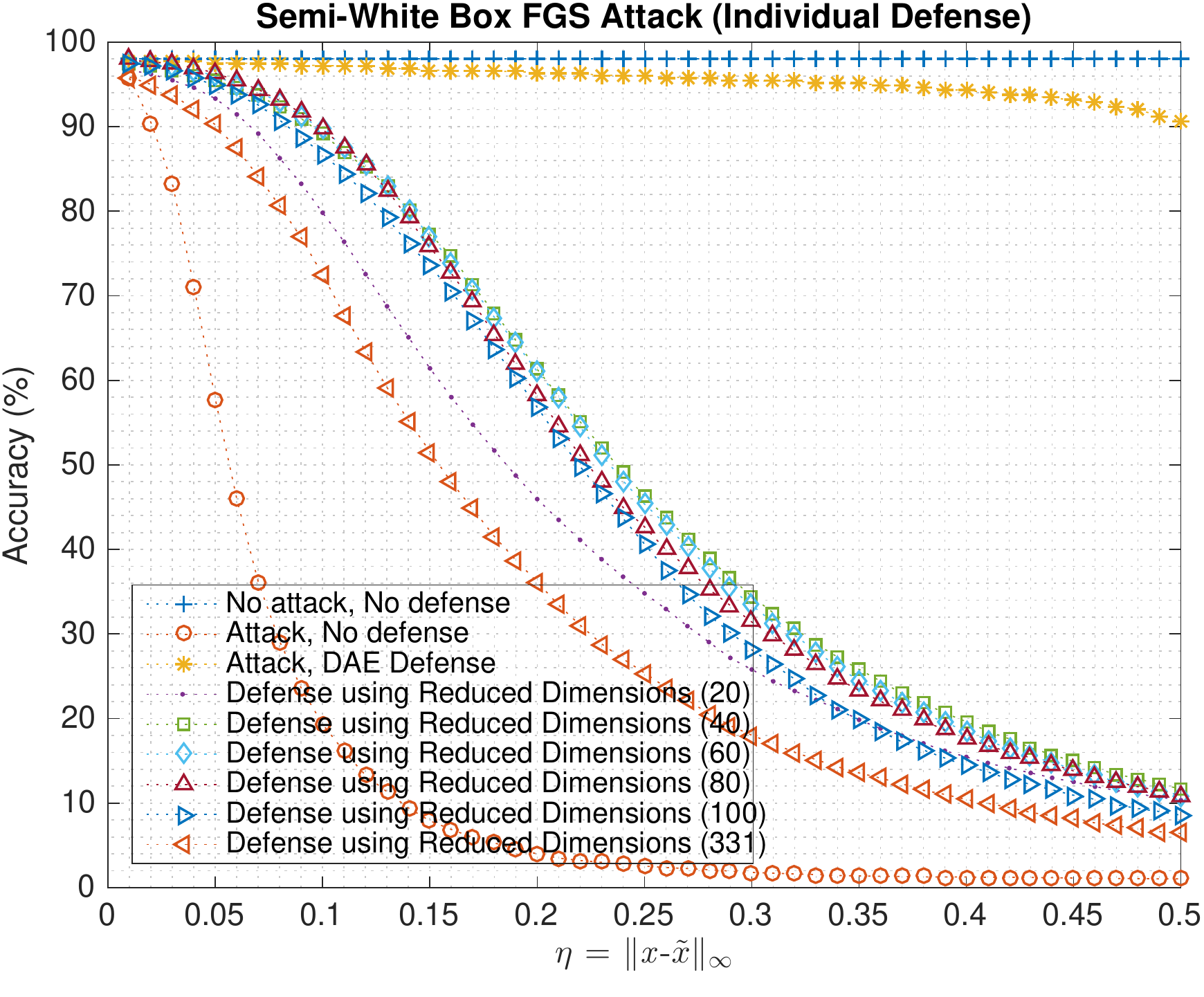}
	\caption{Results for Semi-White Box FGS Attack with standalone DAE and Dimensionality Reduction defenses.}
	\label{fig:swbfgs1}
\end{figure}

The FC-100-100-10 classifier yields an initial classification test accuracy, after being trained on a disparate training set, of $98.00\%$, which confirms the robustness of the classifier to clean input data that. However, after injecting each input sample with an $l_\infty$ bounded noise, with a magnitude perturbation of $\eta=0.25$, the same classifier yields an accuracy of $2.55\%$. To combat such an attack, the corrupted data is propagated through the DAE, and its output is then input into the classifier. After using the DAE to defend an attack on the magnitude of $\eta = 0.25$, the classifier is able to achieve a $95.96\%$ classification accuracy. Similarly, as shown in Figure~\ref{fig:swbfgs1}, we see that defending an attack using the DAE improves classification accuracy nearly $30$-fold for all values of $\eta \in [0.00, 0.50]$. Specifically, we are able to achieve an average classification accuracy of $95.55\%$ over all tested noise levels, which is quite significant compared to the average accuracy of $13.76\%$, achieved using no defense.

Using the hidden layer of an autoencoder to reduce the number of dimensions occupied by input samples results to an increase in accuracy against adversarial attacks. For example, injecting a perturbation of $\eta = 0.25$ into the testing data, and then using an autoencoder to extract the $k = 40$ most significant features of the corrupted input for classification by the FC-100-100-10 model, increases the accuracy from $2.55\%$ to $46.29\%$. Furthermore, reducing the dimensionality to $k=40$ results in an average classification accuracy of $51.09\%$ across all tested noise levels, as compared to an average accuracy of $13.76\%$ classification accuracy when the perturbed data is processed without any defense. Therefore, using a reduced dimension of $k = 40$ provides the best dimensionality reduction defense for the semi-white box FGS attack. The specific improvements in accuracy for all tested levels of noise are shown in Figure~\ref{fig:swbfgs1} below.

The series combination of the DAE followed by dimensionality reduction, using $k=80$ reduced dimensions, provides the best defense compared to using either defense independently. The cascaded defense in which the the denoised data is reduced to $80$ dimension yields an average classification accuracy of $96.27\%$, which is greater than both the $95.55\%$ average classification accuracy, achieved by using only the DAE, and the $46.29\%$ average classification accuracy, achieved by the most robust dimensionality reduction defense of $k=40$. Figure~\ref{fig:swbfgs2} below shows the accuracy of each cascaded defense for all tested noise levels.
\begin{figure}[htb]
	\includegraphics[width=\linewidth]{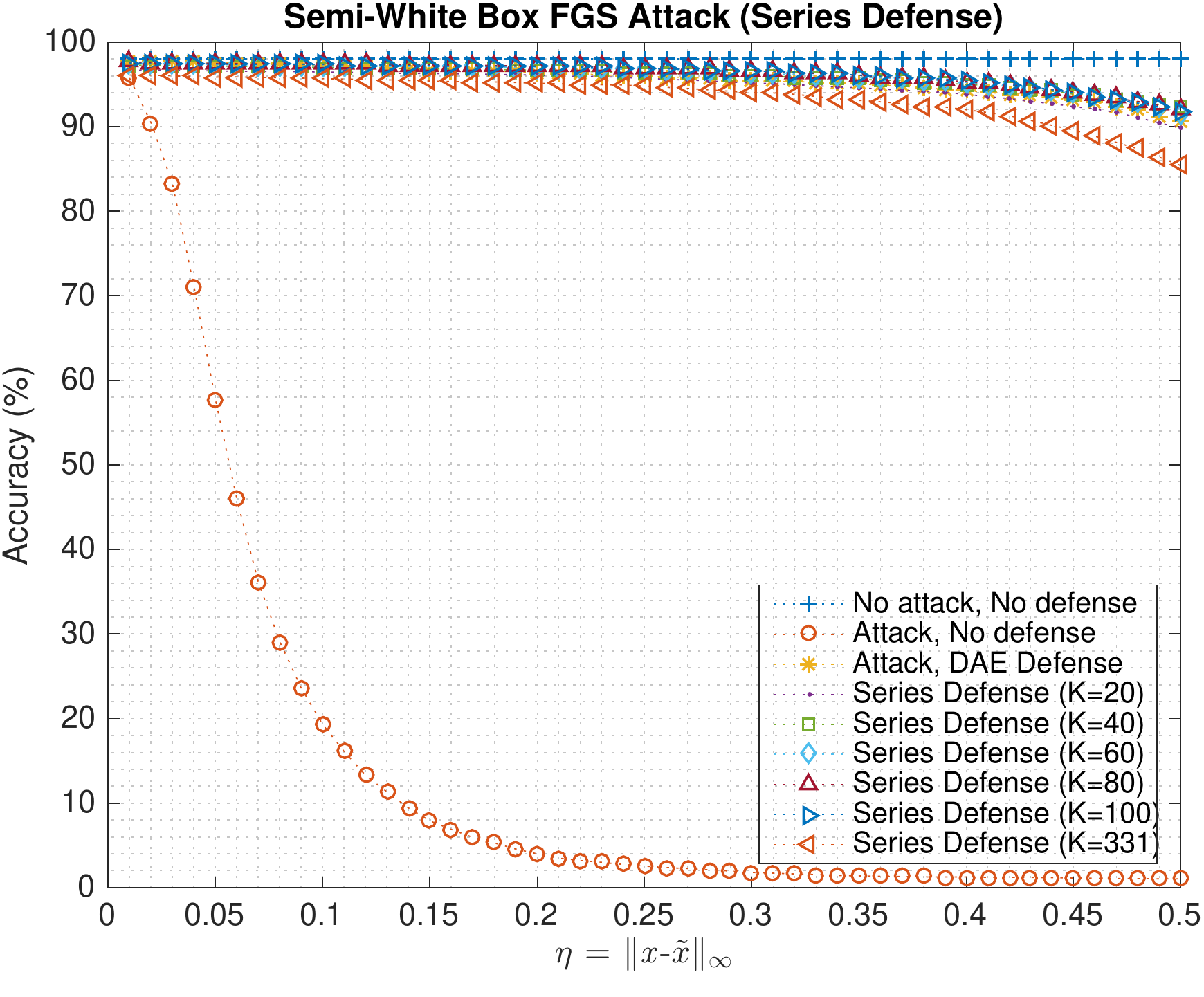}
	\caption{Results for Semi-White Box FGS Attack with a cascaded DAE and Dimensionality Reduction defense.}
	\label{fig:swbfgs2}
\end{figure}
\subsection{Semi-White Box FG Attack}\label{sec:swbfg}
The injection of adversarial noise bounded under the $l_2$ constraint into testing data, without deploying any defense, results in an average accuracy of $27.79\%$ for $\epsilon \in [0, 3.5]$. However, using the DAE trained to combat $l_\infty$ bounded attacks as a defense, the FC-100-100-10 is able to achieve an average accuracy of $85.30\%$ over the same noise range. Figure~\ref{fig:swbfg1} below shows the specific accuracy attained for each noise level before and after introducing the corruption without employing a defense, as well as the classification accuracy of each tested noise level after using the DAE as a defense and then using the FC-100-100-10 classifier model.

Deploying dimensionality reduction as a defense also proves to increase accuracy. Reducing the dataset to $k=80$ dimensions results in the most robust dimensionality reduction classification by increasing the average accuracy from $27.79\%$ to $76.83\%$ across the entire tested noise range. However, using a reduced dimension of $k=40$ and $k=60$ results in an average classification accuracy of $76.03\%$ and $76.48\%$, respectively. Using a lower dimension of $k=20$ yields an average accuracy of $69.12\%$, indicating that relevant features of the data are lost at high levels of compression. Figure~\ref{fig:swbfg1} below shows specific accuracies for each tested noise level for different numbers of reduced dimensions.
\begin{figure}[htb]
	\includegraphics[width=\linewidth]{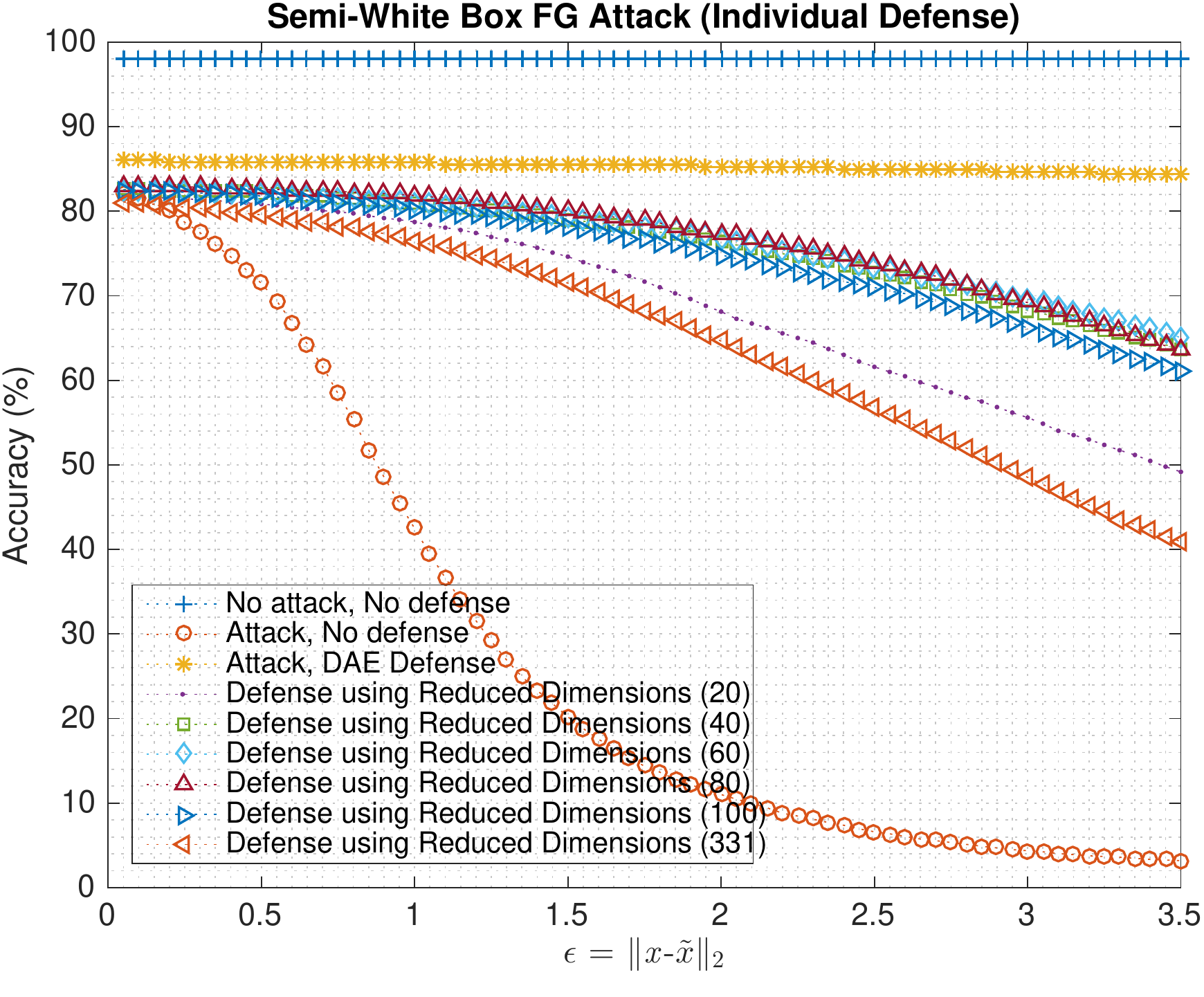}
	\caption{Results for Semi-White Box FG Attack with standalone DAE and Dimensionality Reduction defenses.}
	\label{fig:swbfg1}
\end{figure}
The cascaded architecture, on average, does not outperform the stand-alone DAE defense for any of the experimented reduced dimensions. The cascaded defenses in which $k=40$, $k=60$, and $k=80$ produce an average classification accuracy of $84.68\%$, $85.23\%$, and $85.10\%$, respectively, whereas using the DAE alone yields an average classification accuracy of $85.30\%$. As shown in Figure~\ref{fig:swbfg2} below, the aforementioned reduced dimensions improve the accuracy when used in a cascade, as compared to only the reduced dimensions alone, but the cascaded defense is never able to achieve greater robustness than merely using the DAE as defense.
\begin{figure}[htb]
	\includegraphics[width=\linewidth]{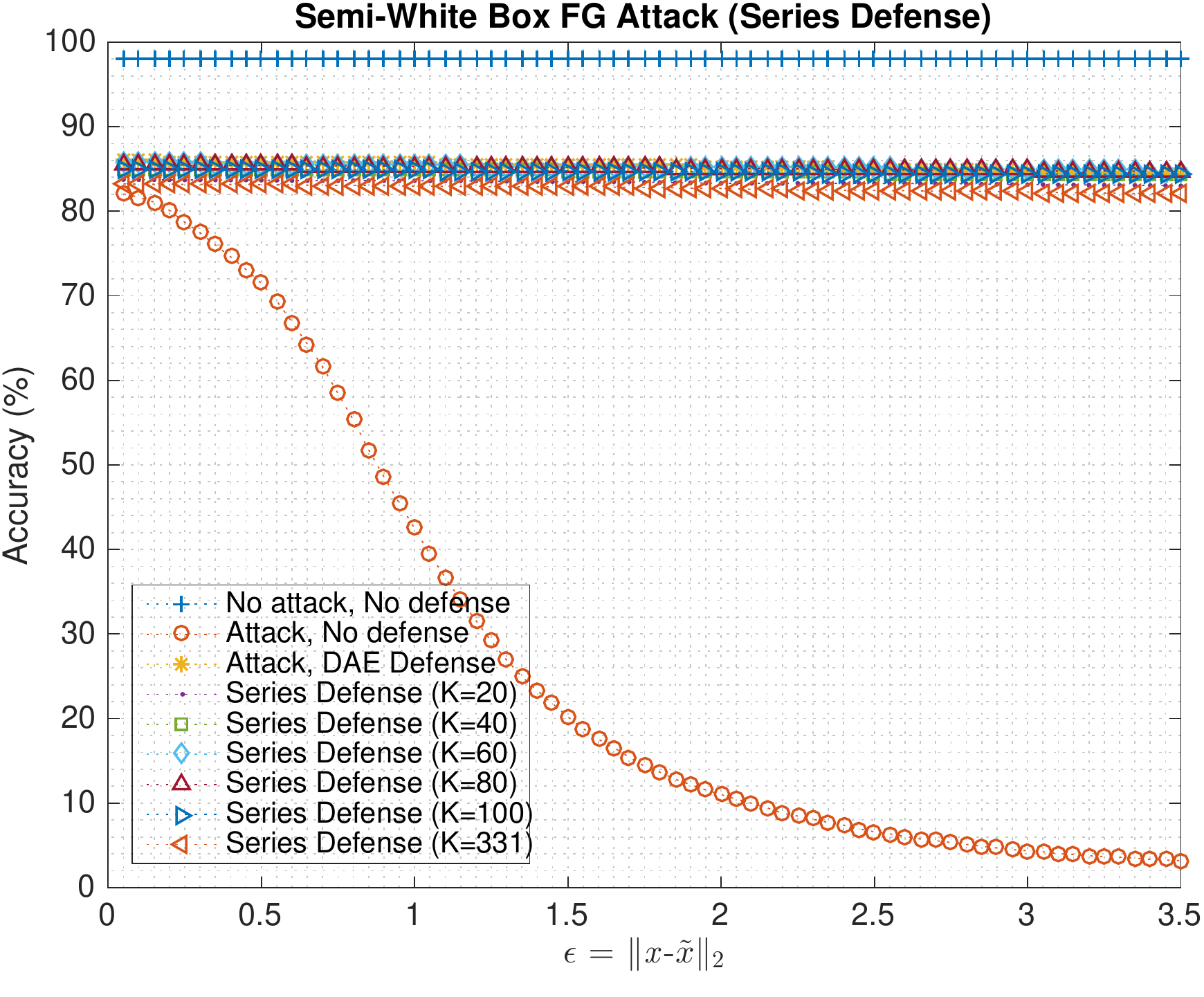}
	\caption{Results for Semi-White Box FG Attack with a cascaded DAE and Dimensionality Reduction defense.}
	\label{fig:swbfg2}
\end{figure}
\subsection{Black Box FGS Attack}\label{sec:bbfgs}
\begin{figure}[htb]
	\includegraphics[width=\linewidth]{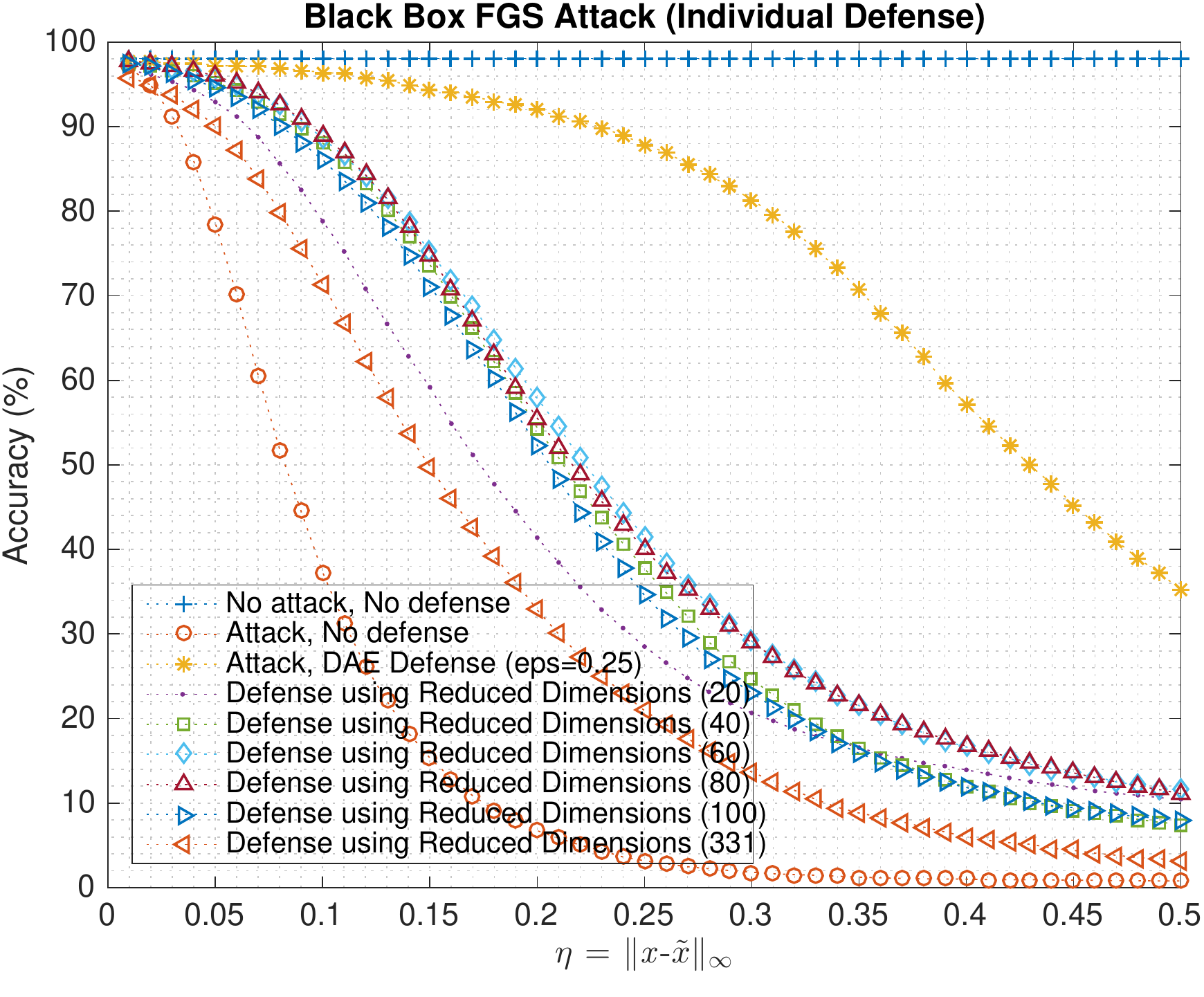}
	\caption{Results for Black Box FGS Attack with standalone DAE and Dimensionality Reduction defenses.}
	\label{fig:bbfgs1}
\end{figure}

As mentioned above, the initial testing accuracy of the FC-100-100-10 model is $98.05\%$, which confirms the robustness of the classifier after it has been trained with similar, but disparate, training data. After the attacker has generated an independent model, corresponding to an FC-200-200-100-10 architecture, each input is attacked according to~\eqref{eq:fgs}, where $J()$ corresponds to the cost function of the attacker’s model. The corrupted inputs are then passed into the defender’s classifier model. The average accuracy, without a defense, is $18.54\%$ over $\eta \in [0.00, 0.50]$. After deploying the stand-alone DAE defense, the defender’s model achieves an increased average accuracy of $78.37\%$ over the same noise range. Figure~\ref{fig:bbfgs1} shows the accuracy achieved for each tested noise level with and without the DAE defense.

The best dimensionality reduction defense, which is achieved using $k=60$, results in an average accuracy of $48.62\%$, which is greater than the average accuracy of $18.54\%$ obtained with no defense. Furthermore, using $k=80$ produces an average accuracy of $48.20\%$. We observe that using $K>80$ allows the reduced dimensional representation to retain more noise than necessary, reducing the average accuracy whereas $k<40$ forces the representation to extort relevant features. This trend can be seen in Figure~\ref{fig:bbfgs1}, which shows the accuracy of different noise levels for each reduced dimensional representation of the data.

The cascaded architecture defense using $k=60$ improves the average accuracy from $18.54\%$ to $79.88\%$. Furthermore, using $k=20$ and $k=40$ results in average accuracies of $79.59\%$ and $79.33\%$, respectively. Each of these cascaded architectures outperforms the DAE as a stand-alone defense, which produces an average accuracy of $78.37\%$ across the entire tested noise range. This phenomenon does not contradict earlier observations about compression reducing accuracy due to the extortion of relevant data features as this trend would still be prevalent for further compressions of the data ($k<20$). Figure~\ref{fig:bbfgs2} below shows the accuracy of each tested noise level for the cascaded defense consisting of different compression levels. 

\begin{figure}[htb]
	\includegraphics[width=\linewidth]{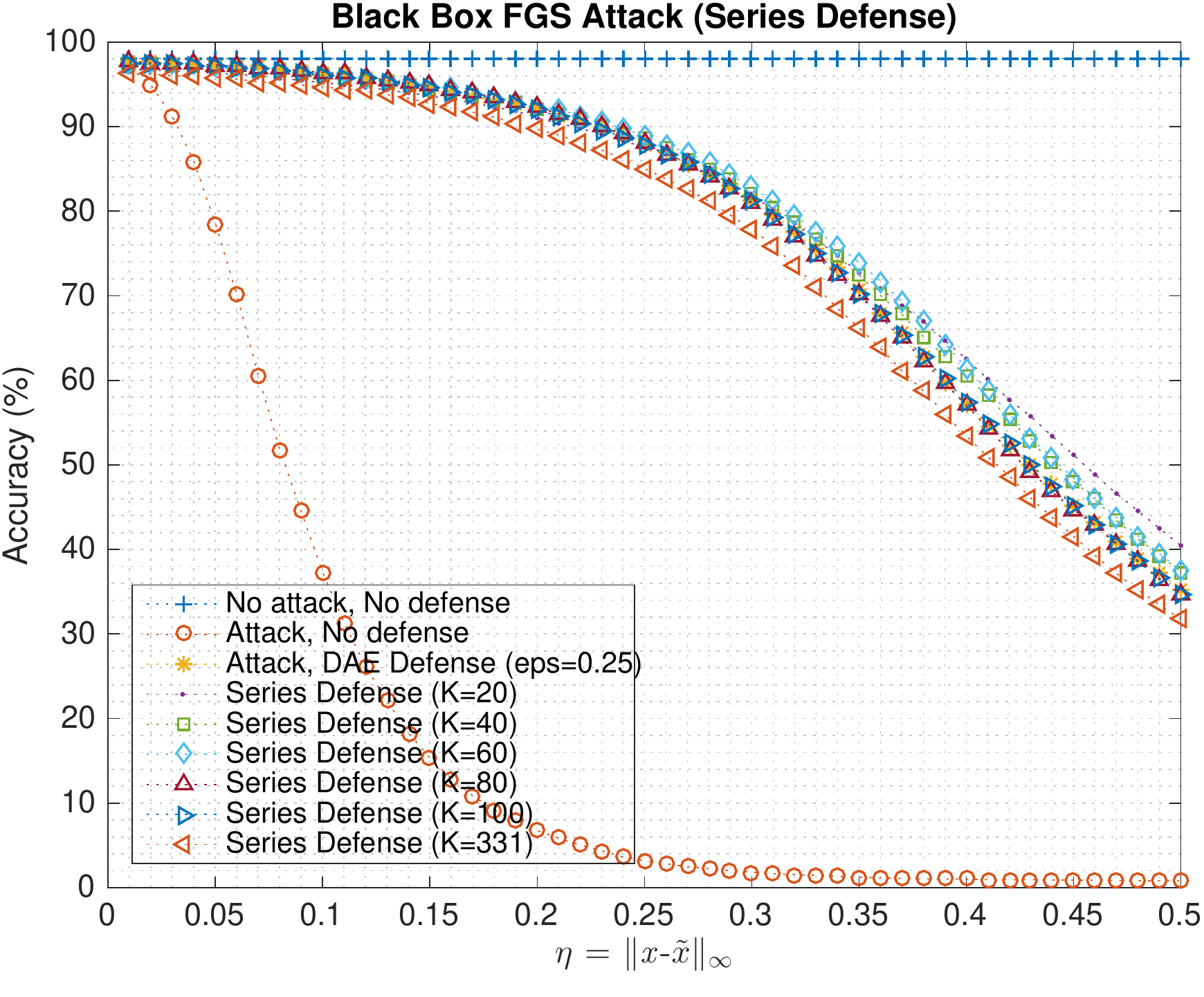}
	\caption{Results for Black Box FGS Attack with a cascaded DAE and Dimensionality Reduction defense.}
	\label{fig:bbfgs2}
\end{figure}
\subsection{Black Box FG Attack}\label{sec:bbfg}
FC-100-100-10 classifier in this scenario achieves an accuracy of $97.88\%$ on the MNIST test data with neither attack nor defense. Introducing an attack based on a disparate FC-200-200-100-10 architecture reduces the average accuracy to $31.08\%$ for $\epsilon \in [0, 3.5]$. Deploying the DAE defense, which was trained to combat $l_\infty$ bounded attacks generated from the FC-100-100-10 classifier, improves the average accuracy to $69.14\%$. Figure~\ref{fig:bbfg1} below shows the attained accuracies at each tested noise level with and without the defense. The trend confirms that the DAE can effectively improve accuracy in a mismatched architecture setting.

Using dimensionality reduction as a stand-alone defense proves to effectively increase the average accuracy in this setting. The most robust defense is achieved using $k=40$, which leads to an average accuracy of $60.48\%$ over $\epsilon \in [0, 3.5]$. Furthermore, using $k=60$ results in an average accuracy of $59.83\%$, whereas using $k=20$ results in an average accuracy of $56.35\%$. The loss of accuracy experienced by the increase or decrease in the number of dimensions further confirms the retention of excess noise or extortion of relevant features, respectively, for a reduced dimensional representation of the input. Figure~\ref{fig:bbfg1} shows the accuracy at each tested noise level for different reduced dimensional subspaces as stand-alone defenses.

\begin{figure}[htb]
	\includegraphics[width=\linewidth]{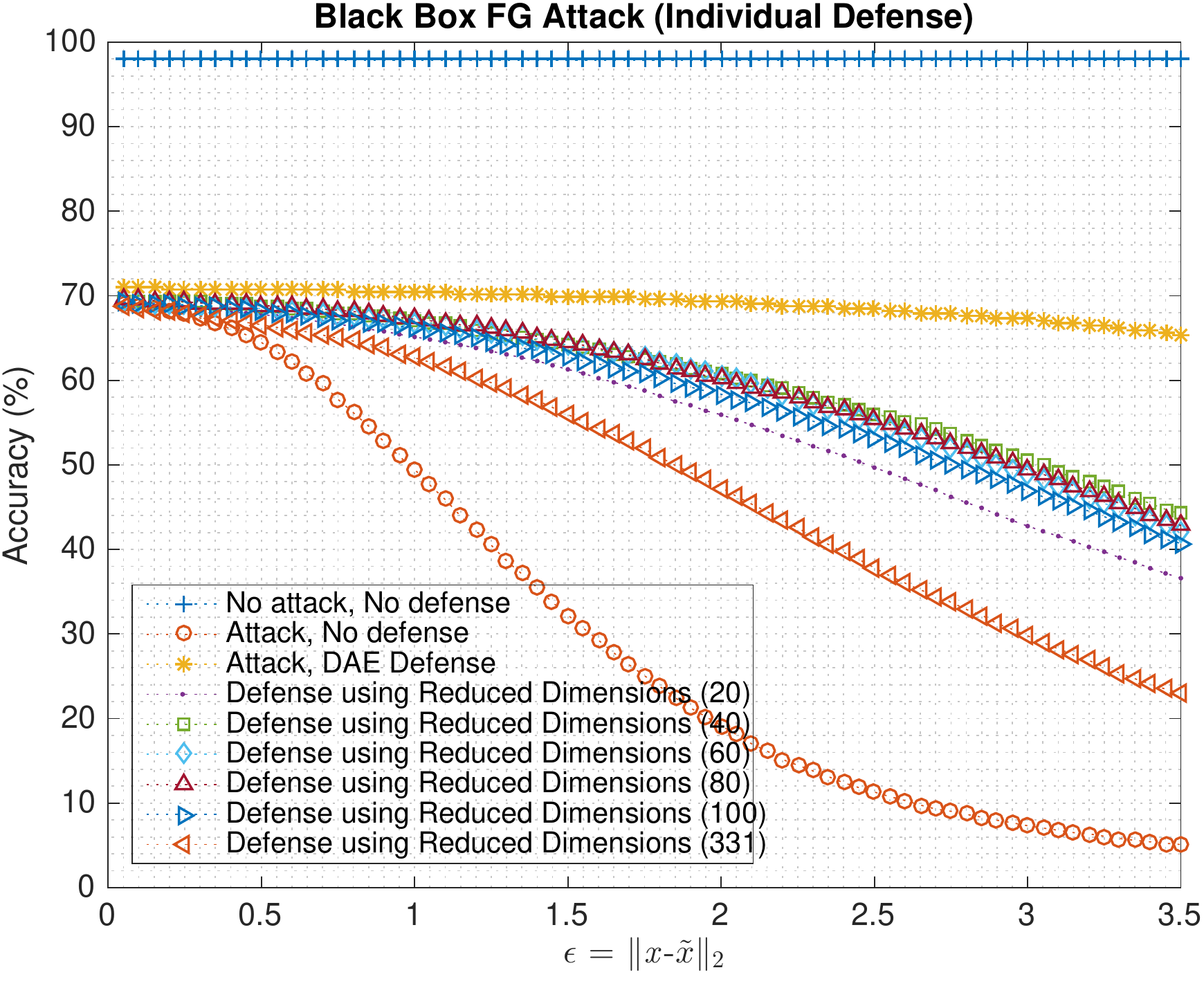}
	\caption{Results for Black Box FG Attack with standalone DAE and Dimensionality Reduction defenses.}
	\label{fig:bbfg1}
\end{figure}

\begin{figure}[htb]
	\includegraphics[width=\linewidth]{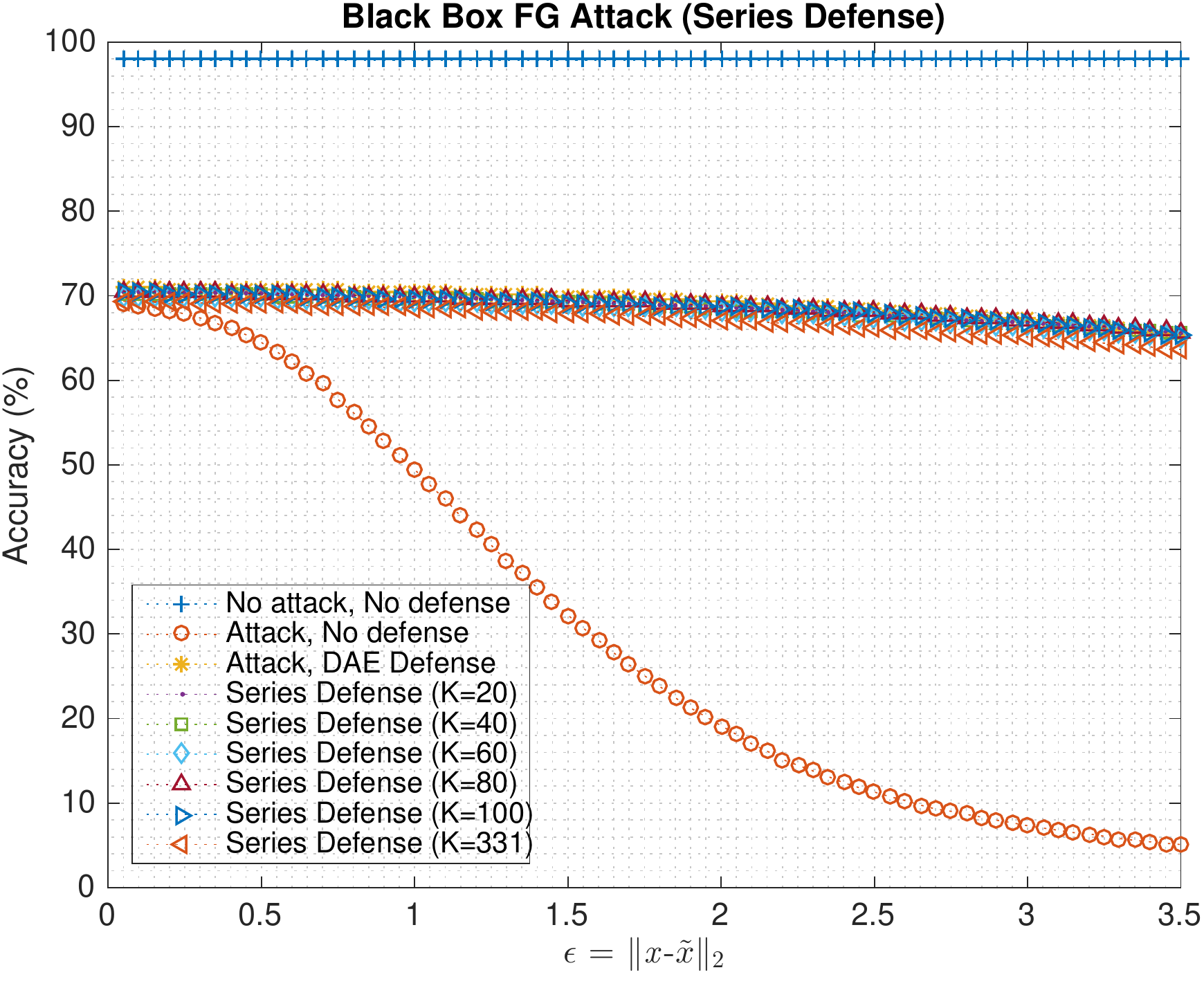}
	\caption{Results for Black Box FG Attack with a cascaded DAE and Dimensionality Reduction defense.}
	\label{fig:bbfg2}
\end{figure}

Similar to the semi-white box FG attack, the cascaded defense does not, on average, outperform the stand-alone DAE for any of the experimented reduced dimensions. The most robust cascaded defense is the architecture in which the DAE is followed by a reduced dimension of $k=80$, results in an average accuracy of $68.78\%$ across the tested noise range. Also, using a reduced dimension of $k=60$ and $k=100$ results in average accuracies of $68.22\%$ and $68.56\%$, respectively, whereas the stand-alone DAE yields an average accuracy of $69.14\%$ across the same noise range. Figure~\ref{fig:bbfg2} shows the accuracy of the cascaded architectures at each experimented noise value and reveals that none of the cascaded defenses outperforms the stand-alone DAE at any tested level of noise.

\section{Concluding Remarks}
The effectiveness of both the DAE and dimensionality reduction using an autoencoder can be attributed to autoencoders’ abilities to effectively compress data. Throughout these experiments, our autoencoder architectures were designed to compress the input into a desired, reduced dimensional representation, and then reconstruct an approximately equivalent version of the input at the output. 
Our experiments confirmed that DAEs trained to combat $l_\infty$ bounded attacks are effective in denoising $l_2$ bounded attacks due to the architecture being trained to accurately map larger corruptions, compared to $l_2$ bounded attacks, to the original data value. Lastly, our experiments confirmed that a DAE is capable of denoising perturbations on our entire tested noise range, even though we are training it using only perturbations on the scale of $\eta = 0.25$. This reveals that, with relatively little training data, the defender can effectively denoise higher and lower perturbation levels than what its training set consisted of. 

In the case of applying an autoencoder for dimensionality reduction, for both the stand-alone and cascaded defense, we observe that the value of $k$, which produces the most robust defense, is not standard across attack environments. This phenomenon is due to several factors including, but not limited to, the training data, the classifier architecture, and the training parameters. The exploration of fine-tuning an optimal reduced dimensional representation was outside the scope of this work, but it will lead the direction of future work. As a preliminary attempt to extend this work, we compared our autoencoder-based dimensionality reduction method with Principal Component Analysis (PCA), and the initial results strongly suggest that using deeper autoencoders for dimensionality reduction and fine tuning other network parameters, as discussed above, could result in higher defense accuracies than using PCA alone. As a result, we believe that the potential of the proposed cascaded defense would be more significant in such scenarios. 



Overall, our experiments revealed novel defense mechanisms capable of combatting adversarial attacks on neural network classifiers. The DAE can deliver a significant gain in accuracy for corrupted data, as compared to using no defense, as well as retain high accuracy levels when processing benign inputs. Further, this paper showed that shallow autoencoder networks used to reduce the dimensionality of the input is another effective defense against adversarial attacks
Finally, our cascaded architecture, consisting of the DAE followed by dimensionality reduction for various values, proved to be more effective than either stand-alone defense when the attacker introduced the FGS attack, but it showed to be slightly less robust than using the DAE alone when inducing the FG attack. 


\bibliographystyle{IEEEtran}

\end{document}